# Learning Object Arrangements in 3D Scenes using Human Context

**Yun Jiang**                                                    YUNJIANG@CS.CORNELL.EDU
**Marcus Lim**                                                        MKL65@CORNELL.EDU
**Ashutosh Saxena**                                              ASAXENA@CS.CORNELL.EDU
Department of Computer Science, Cornell University, Ithaca, NY 14850 USA

## Abstract

We consider the problem of learning object arrangements in a 3D scene. The key idea here is to learn how objects relate to human poses based on their affordances, ease of use and reachability. In contrast to modeling object-object relationships, modeling human-object relationships scales linearly in the number of objects. We design appropriate density functions based on 3D spatial features to capture this. We learn the distribution of human poses in a scene using a variant of the Dirichlet process mixture model that allows sharing of the density function parameters across the same object types. Then we can reason about arrangements of the objects in the room based on these meaningful human poses. In our extensive experiments on 20 different rooms with a total of 47 objects, our algorithm predicted correct placements with an average error of 1.6 meters from ground truth. In arranging five real scenes, it received a score of 4.3/5 compared to 3.7 for the best baseline method.

## 1. Introduction

*"We bear in mind that the object being worked on is going to be ridden in, sat upon, looked at, talked into, activated, operated, or in some other way used by people individually or en masse."* Dreyfuss (1955).

In fact, in human environments, arrangements of objects are often governed by their affordances. For example, the objects in the 3D scene in Fig. 1 are arranged in a particular configuration because they are meant to be used by humans for activities such as watching TV, working on laptop, etc. In another example, a keyboard is found below a monitor because it needs to be reachable by hand when the monitor is in sight. Learning such affordances for reasoning about the objects' arrangements would be useful in several areas of scene understanding (e.g., Koppula et al., 2011) or assistive robots (e.g., Jiang et al., 2012a).

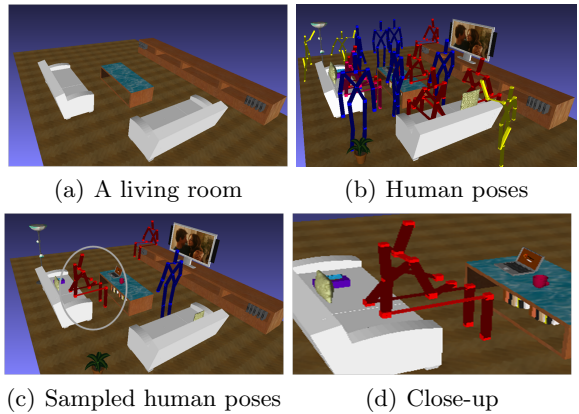

(a) A living room    (b) Human poses

(c) Sampled human poses    (d) Close-up

*Figure 1:* Given a scene (a), there are many possible human poses in it (b). Because of the human-object relationship, only a few are meaningful (c). Our goal is to learn object arrangements by modeling human poses and how objects relate to human poses through their affordances (d).

In this work, we take an unsupervised learning approach to this task. Given a collection of 3D scenes containing objects, we learn how an object relates to a human pose for an activity by defining a parameterized density function. While at the first blush, introducing human poses may seem to complicate the model, it actually simplifies it by making it more parsimonious. The reason for this is that the set of relevant human poses is far smaller than the collection of all objects. If we learn object to object relationships instead of learning human pose to object relationships, the complexity of both model and computation grows quadratically in the number of objects. Human pose, which is the underlying factor connecting the objects to each other and the scene, provides a more parsimonious model for the arrangement of objects in a scene.

In order to model the object arrangements, we use a Dirichlet process (DP) mixture model for defining the joint distribution of human poses and objects. We treat each human pose as a mixture component that models the distribution of objects, and we use DP to determine which human pose the object is generated from. This model allows different objects to be used by the same human pose (e.g., using a monitor, keyboard and mouse at the same time), with different parameters in the density function. However, same objects will have same parameters across different human poses. This requires that we formulate our learning method as maximum likelihood estimation based on human poses sampled using DP. This variant of DP mixture model allows the same parameters to be shared by the instances from the same category across different mixture components.

In this work, we specifically consider learning object arrangements with a robotic application in mind—a personal robot arranging a disorganized house. Jiang et al. (2012a;b) considered a similar problem. However ignoring the role of humans often led to unreasonable placements. Inference in our model naturally follows how we usually organize a room: from a given room, we first infer possible workspaces such as sitting on the chair and standing by the kitchen shelf; then we reason about how to place the objects for potential activities, such as placing a laptop on the desk facing the chair.

In our experiments, we collected a large dataset comprised of different scenes from three categories: kitchen, living room and office. Each scene was manually labeled by three to five subjects, who determined where and how to place the given objects. In total, 19 different types of object are placed. The experimental results demonstrate that our methods can find reasonable locations and elevations for objects in most cases. When evaluated by human subjects on real point-clouds, the average score is 4.3 out of 5, compared to 3.7 from the best baseline.

## 2. Related Work

Most previous work in vision and robotics has considered the problem of scene understanding and its application to robotics using 2D images. With recent inexpensive RGB-D sensors, it is possible to obtain full 3D point-clouds and therefore reason about human poses in 3D. There are also some previous works that consider human pose and activity recognition, and we describe them below.

**Scene understanding.** A number of works propose approaches to capture the relations between different parts of the object (Felzenszwalb et al., 2008) and between different objects in 2D images (Heitz & Koller, 2008). Some works extract 3D scene geometry from a single image for object detection (e.g., Saxena et al., 2005; Hoiem et al., 2006; Heitz et al., 2008; Lee et al., 2010; Li et al., 2010). The recent availability of RGB-D sensors provides more precise geometry of indoor scenes and enables capturing stronger context among objects (Koppula et al., 2011). The goal of these works is to find and label existing objects in a scene, while our goal is to infer arrangement of objects in the scene. Furthermore, these methods focus on learning object-object relationships and therefore do not scale well with large number of objects. By contrast, our work models the role of human poses as the underlying reason for object arrangements.

**Related applications.** There is little work in robotic placing and arrangement of objects. Edsinger & Kemp (2006) and Schuster et al. (2010) focused on finding flat clutter-free areas where an object could be placed, but did not model any form of semantic context for meaningful placing locations. Fisher et al. (2011) considered the problem of finding the most visually relevant object to be placed in a given location. Jain et al. (2009) considered symbolic planning for arranging objects such as setting a dinner table. Their work does not address finding desired object locations and is complementary to ours. Jiang et al. (2012b;a) employed 3D stability and geometric features to find stable and preferred placements. While geometry is an important hint for context, it alone cannot tell the difference between a TV facing towards a couch and facing towards the wall. In order to learn such relations, we need to consider the role of human poses in meaningful object arrangements.

**Human pose estimation and activity detection.** Estimating and understanding human pose in both static images and videos has attracted great attention in the computer vision community (e.g., Lee & Cohen, 2006; Ly et al., 2012). Some other work uses human poses to facilitate high level understanding, such as human activity detection (Sung et al., 2012). While these studies try to abstract human poses when human beings are present, there have been some work making use of imaginary human poses to detect objects (Grabner et al., 2011) and to infer human workspaces (Gupta et al., 2011). Our work, inspired by this viewpoint, uses human poses to detect the affordances of the object/scene to gain a deep understanding of the human environments. We take it further to inferring object arrangements based on the learned object affordances.

## 3. Overview

We first generate a set of human poses in the scene based on certain criteria (such as reachability or usage with existing objects). We then use these human poses to estimate placements for new objects. It is the hu-man poses that link objects together. For example, a monitor on the desk could generate a sitting skeleton in front of it. Then the sitting skeleton could further suggest to place a mouse close to the hand and therefore at the edge of the desk. Note that objects are related by this means naturally. The interaction affects the human poses and object placements simultaneously. There are two components to this as described below:

**Human access and usability cost.** One of the objectives while arranging a room is to make it convenient for humans to use objects. For example, people usually prefer frequently-used objects to be placed on a table top, rather than on the floor, to reduce the access effort. A television is often facing an open area than the wall to increase its usability. Also, objects related to the same human activity are often grouped together, such as dishware and utensils on a dining table. In order to capture this in our model, we define a potential function over human poses and objects, based on a collection of features modeling human-object interaction, such as their distance, relative orientation and activity matching (see Section 4.1).

**Sampling of human poses.** While there could be innumerable potential human poses in a scene, only a few of them are meaningful and relevant. For example, acrobatic or dancing poses are possible but rarely appear in a room. Certain human poses are more likely than others (and also more relevant for object interactions/usage), such as sitting and standing. The potential poses are also restricted by the layout of the scene, because of potential collisions, usage interactions with existing objects in the room, and their kinematic cost. For instance, in a room such as Fig. 1, a standing pose facing against the corner is less important because it does not connect with any object. We address this by sampling human poses according to the potential function using a DP (Section 4.2).

Certain objects have similar purposes and are commonly placed together, such as a PC setup of monitor/keyboard/mouse, a dishware set, or a TV and a remote control. This is because the human poses and object classes are linked via an activity. For example, standing pose and dishware/utensils are related through cooking activity. In our model, we use human activities to match relevant human poses and objects so that such objects would share similar human pose and activity and hence be placed together.

## 4. Algorithm

In this section, we first define a potential function to quantify the relationships between human poses and object placements, and then present a DP-based algorithm infer poses and placements together. During training, we are given the objects in the scene, and our goal is to learn the distribution of human poses and the parameters of the potential function.

Formally, given a new scene, denoted by $E$, which may contain some already placed objects $\mathcal{G} = \{G_1, G_2, \ldots, G_n\}$, our task is to arrange more objects in the scene, $\mathcal{O} = \{O_1, \ldots, O_m\}$. Each object $G_i$ or $O_j$ is specified by its type, location and orientation. We also define the set of all possible human poses as $\mathcal{H}$. A human pose is specified by its joint locations and activity (see Section 4.4 for more details).

### 4.1. Potential Function

To describe the human-object relationship, we define the potential function between a human pose $H$ and an object $O$ (or $G$) as

$$\Psi(O, H; \Theta) = \Psi_{\text{dist}} \Psi_{\text{rel}} \Psi_{\text{ori}} \Psi_{\text{h}} \Psi_{\text{OA}} \Psi_{\text{PA}}. \quad (1)$$

It consists of several terms, as described below:

*Distance preference.* Some objects are preferred to be at a certain distance from humans, such as a TV or a laptop. This preference, encoded as $\Psi_{\text{dist}}(O, H)$, includes how far the object should be, from what joint of the human skeleton, and how strong this bias is. The Euclidean distance between $O$ and a designated joint of $H$ follows log-normal distribution.

*Relative angular preference.* There is a preference for objects to be located at a certain angle with respect to human poses. For example, people will sit in front of a laptop, but prefer the mouse to be on their right (or left). We use a von Mises distribution for $\Psi_{\text{rel}}$.

*Orientation preference.* There is a preference for objects to be oriented at a certain angle with respect to the human pose (e.g., a monitor should also be facing towards the skeleton when located in front of the skeleton). Similarly to relative angular preference, $\Psi_{\text{ori}}$ is also a von Mises distribution over the difference between orientations of $O$ and $H$.

*Height preference.* $\Psi_{\text{h}}$ is a Gaussian distribution of the object's relative height to a human pose.

*Object-activity and pose-activity preference.* Different objects can be related to different sets of activities, and so can human poses. We use two terms to represent the relationship between object and activity ($\Psi_{\text{OA}}$) and between human pose and activity ($\Psi_{\text{PA}}$). We represent them as probability tables.

The potential function relates an object $O$ to a human pose $H$ using the object-specific parameters $\Theta$. These parameters consist of parameters from each of the terms above, such as the shape and log-scale parameter of the log-normal distribution in $\Psi_{\text{dist}}$ and the mean and concentration parameter of the von Mises distribution in $\Psi_{rel}$ and $\Psi_{ori}$. These parameters $\Theta$ are shared by objects from the same category. We now present how to sample $H$ and learn $\Theta$ using DP.

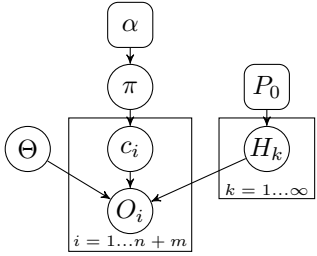

*Figure 2:* The DP mixture model for generating placements $O_i$ for each object. $\alpha$ is the concentration hyper-parameter of DP. $\pi$ is the resulting distribution from stick-breaking process, and $c_i$ is the label for the placement $O_i$, denoting which human pose is selected for $O_i$. $H_k$ are the samples of human poses from $P_0$. $\Theta$, which is not in DP, is the object-specific parameter of the potential function.

## 4.2. Dirichlet Process Mixture Model

In this section, we describe our formulation of the object arrangements as a DP mixture model. The inference in our model is similar to the standard DP (see Teh (2010) and Neal (2000) for an overview). However our learning method differs from it, as described in Section 4.3.

Our particular problem can be viewed as a generative process (shown in Fig. 2), we treat each human pose as a mixture component that models the distribution of objects, and we use DP to determine which human pose the object is generated from. More formally, we first draw the prior distribution $P(H)$ from $DP(P_0, \alpha)$, where $P_0$ is the base distribution and $\alpha$ is the concentration parameter. Then a human pose $H$ is drawn from $P(H)$. Finally, the object's feature (i.e., location and orientation) is selected following $\Psi(O, H; \Theta)$.

In this work, to estimate the distribution of $\mathcal{O}$ generated by DP, we adopt the method of Gibbs sampling with auxiliary parameters in Neal (2000). In testing, this method iteratively samples the assignments $\mathbf{c}$, human poses $\mathcal{H}$ and placements $\mathcal{O}$. Suppose there are $K$ distinct $c_i$ for $i = 1, \ldots, n + m$, where $n + m$ is the total number of objects in the scene. In other words, we have $K$ human poses. We first augment the number of human poses to $K + z$ by sampling $z$ auxiliary human poses from the base distribution $P_0$. Then, for every object in either $\mathcal{O}$ or $\mathcal{G}$, we sample $c_i$ as,

$$c_i = c | c_{-i}, O_i, \mathcal{H} \propto \begin{cases} \frac{n_{-i,c}}{n+m-1+\alpha} \Psi(O_i, H_c; \Theta_i) & n_{-i,c} \geq 0, \\ \frac{\alpha/z}{n+m-1+\alpha} \Psi(O_i, H_c; \Theta_i) & \text{otherwise} \end{cases}$$

where $c_{-i}$ denotes other assignments in $\mathbf{c}$ except $c_i$, and $n_{-i,c}$ represents the number of assignments in $c_{-i}$ that take the value $c$. This equation shows that the distribution of $c_i$ will be selected according to the potential score and the 'popularity' of a human pose ($n_{-i,c}$). It also depends on the concentration parameter $\alpha$ that controls the probability of selecting a new pose.

After the assignments are selected, $\mathcal{H}$ and $\mathcal{O}$ are selected based on their posterior:

$$H_k | \{O_i | c_i = k\} \quad \propto \quad \prod_{i:c_i=k} \Psi(O_i, H_k; \Theta_i) P_0(H_k), \quad (2)$$

$$O_i | c_i, \Theta, H_{c_i} \quad \propto \quad \Psi(O_i, H_{c_i}; \Theta_i). \quad (3)$$

By this means, we obtain a collection of sampled objects, $\{\mathcal{O}^1, \ldots, \mathcal{O}^s\}$. Since they are drawn from the distribution of $\mathcal{O}$, we approximate the distribution by counting the samples near it, i.e., $O_i \approx \frac{1}{s} \sum_{j=1}^{s} I\{O_i^j \in \Omega(O_i)\}$, where $I$ is the indicator function and $\Omega$ represents a small neighborhood around this placement.

## 4.3. Learning Object-Specific Parameters

During training, we are given scenes with placed objects $\mathcal{G}$ and our goal is to learn object-specific parameters $\Theta$ that maximize the potential over all the scenes with latent human poses. Since $\Theta$ is invariant to the scene and human pose, we do not sample them in DP, but rather learn them using maximum likelihood estimation (MLE).

In detail, we use a DP to sample poses $H^1, \ldots, H^s$ as our observations. The optimal $\Theta$ is then given by,

$$\Theta^* = \arg\max_{\Theta} \sum_{scenes} \sum_{j=1}^{s} \sum_{i=1}^{n} \log \Psi(G_i, H_{c_i}^j; \Theta) \quad (4)$$

We can optimize $\Theta_O$ independently based on the placements from the category $O$ only. Furthermore, because our potential function is the product of several components in Eq. (1), we can also optimize the parameters in different terms separately. In detail, as $\Psi_{dist}$ and $\Psi_h$ are Gaussian distributions, the posterior mean and variance have closed form, and as $\Psi_{rel}$ and $\Psi_{ori}$ are von Mises distribution, their mean $\mu$ and concentration $\kappa$ can be estimated numerically. We iterate between computing $\Theta^*$ using Eq. (4) and sampling $H$ until convergence.

## 4.4. Human Pose and Object Placement Generation

This section describes the set of human poses and object placements, which is used by DP to sample from.

We extract human skeletons from the Kinect RGB-D dataset (Sung et al., 2012), which contains activities performed by four subjects. We then cluster the poses using k-means algorithm giving us six types of skeletons (see Fig. 3). We sample the variants of these skeletons and check for collisions. In addition to the location, every sampled human poses is also assigned

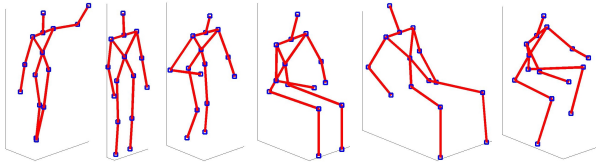

*Figure 3:* Six types of human poses extracted from Kinect 3D data. From left: reaching, standing, leaning forward, sitting upright, sitting reclined and sitting forward.

with an activity, used in $\Psi_{OA}$ and $\Psi_{PA}$ in Eq. (1).[1]

We consider all the surface points of the environment as potential placing locations, and consider eight orientations evenly sampled from 0 to $2\pi$. For each sample, we perform a stability check (if the point has a large enough region to support objects) and collision check (if the bounding boxes of objects overlap).

## 5. Baseline Methods

For comparison in Section 6, we designed a number of baseline algorithms, including a discriminative classifier and a method based on a finite number of mixtures.

**Open-area preference.** Following Schuster et al. (2010), we find a clutter-free area based on distances from already existing objects, and place objects in orientations closest to those in the training set.

**Height preference.** We compute the average height of each object type's placements. This helps in cases such as food or monitors that would be placed on a table (with the height of around 0.5m), while shoes and floor-lights are usually placed on the ground.

**Room - object context.** This method considers the object's relative location in a room. We first normalize the room's size in the training data, and use the average relative location to place the object in testing.

**Object context.** While our goal is to show that we can learn object arrangements only using how they relate to human poses, we also note that object - object context could be a piece of complementary context. As an example, a keyboard is placed in front of the monitor and utensil is often on the side of dishware. We learn object - object context $\Psi_{obj}(O, \mathcal{G})$ as follows. We model the relative location/orientation to place the object as a Gaussian distribution, with parameters extracted from training data. To select the reference object, we compute the variance of the relative placement for every category and choose the one with the smallest variance.

**Discriminative Classifier ('class').** Selecting a good placement for an object can also be treated as a binary classification problem. We build a logistic re-gression classifier for every category with a total of 97 features based on the values used above: the relative distance/orientation to other objects, height, relative XY location and size of its bounding boxes, etc.

**Finite Mixture Model.** We also compare our algorithm with a finite mixture model, where the number of human poses is fixed. Suppose we have $K$ human poses in a scene, denoted by $\mathcal{H} = \{H_1, \ldots, H_K\}$. Similar to the infinite mixture model, given the assignment $c_i \in \{1, \ldots, K\}$, $O_i$ is distributed according to Eq. (3). After marginalizing out $c_i$, we get

$$\mathcal{O}|\mathcal{H}, \Theta \propto \prod_{i=1}^{m} \sum_{c=1}^{K} \Phi(O_i, H_c; \Theta) P(c|\mathcal{H}), \quad (5)$$

where $P(c|\mathcal{H})$ is the probability of choosing $H_c$ among $K$ poses. The inference problem, finding $\mathcal{O}$ that maximizes the potential is solved using an expectation-maximization (EM) algorithm. However, the M-step requires joint optimization over $\mathcal{H}$ and $\mathcal{O}$, which we solve by iteration. In the training, we learn $\Theta$ using MLE (similar to Eq. (4)), i.e.,

$$\max_{\Theta} \sum_{\text{scenes}} \sum_{i=1}^{n} \log \sum_{c_i=1}^{K} \Phi(G_i, H_{c_i}; \Theta) P(c_i|\mathcal{H})$$

This is also estimated using EM with an iterative M-step between $\Theta$ and $\mathcal{H}$.

Compared to our DP approach, a finite mixture model has two major drawbacks: (1) $K$ is pre-defined, which limits the number of human poses varying across rooms with different size and layout; (2) It does not encode the prior of $H$, which is often informative. For instance, the probability of a human pose sitting on a chair is higher than standing up.

**Combining Object Context and Human Context.** We additionally present another algorithm in which we combine the distribution of objects generated through human poses $O \propto \Psi_{\text{human}}(O, H; \Theta)$ with a distribution generated through object - object context $O \propto \Psi_{\text{obj}}(O, \mathcal{G})$ using a mixture model: $O \propto \omega \Psi_{\text{human}}(\cdot) + (1 - \omega) \Psi_{\text{obj}}(\cdot)$. We give a comparison of methods of using object context only, human context only and their combination in our experiments.

## 6. Experiments

In this section, we extensively evaluate our algorithms from three perspectives: (1) robustness across a variety of different scenes and objects; (2) comparison between our method and several baselines, including an approach that models object-wise relationships; (3) different placing scenarios such as placing new objects and placing in an empty room.

**Dataset.** We created a dataset consisting of 20 scenes in three categories: six living rooms, seven kitchens and seven offices. We downloaded the 3D model for each scene from Google 3D Warehouse.[2] All the scenes are commonly seen in the real world and have a large variety in space, layout, furniture, etc. We also gathered a collection of daily objects, such as dishware, books, fruit, lamps, computers, etc, for a total of 19 different types (listed in Table 1). Every room is assigned a set of 10 to 30 objects, and we asked three to five subjects (not associated with the project) to manually label the placements of every object in the scene. A snapshot of our dataset is shown in Fig. 4.

---

[1]In this work, we consider five activities: reading, working, talking, writing and resting.

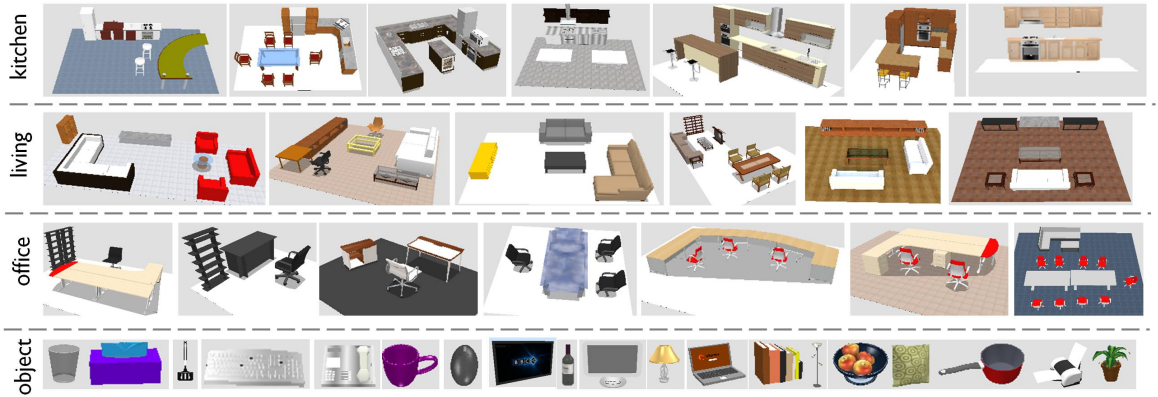

Figure 4: Our dataset contains 20 scenes (7 kitchens, 6 living rooms and 7 offices) and 47 objects from 19 categories that are commonly seen in these scenes.

**Results.** The first experiment was performed on the 20 rooms, with 5-fold cross validation. In each fold, labels of 16 rooms were used for training and the other four rooms for testing, so that the test rooms had never been seen by the algorithm. We created two placing scenarios for test: placing new objects and placing in an empty room. In the first case, for every test room we took out the objects of the type being placed and left other types as given. In the second case, the test rooms had no object in it at all.

We wanted to answer following questions:

*Is the learned density function meaningful?* We visualize some learned density function in Fig. 5, where we put a skeleton facing right and centered at a room of 10m×10m. It shows a 2D distribution of placing different objects relative to the human pose. We can see that TV prefers some distance to the human position, and it has a narrow range of relative orientation, unlike the remote and decoration. The mouse, laptop and dishware all prefer a smaller distance, but have different preferred relative orientations.

*Do sampled human poses and object placements reflect meaningful distributions?* Fig. 6 shows sampled human poses and object placements for some scenes.

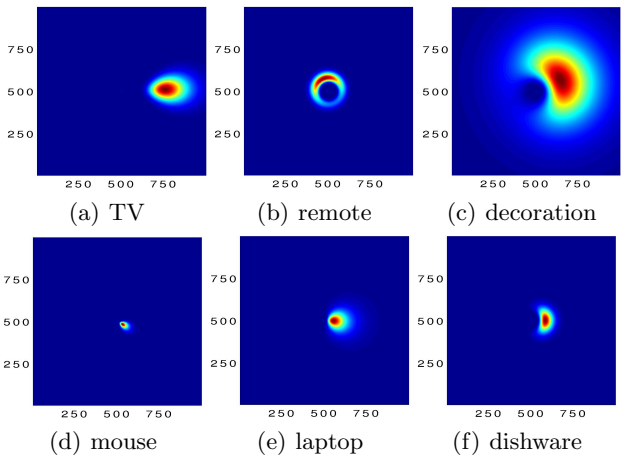

Figure 5: Given a pose in a room with 10m× 10m, at the coordinate of (500, 500) facing right, distributions of different objects according to the learned density function.

When placing a monitor in the first scene, the existing objects caused human poses to be sampled at the corner side of the desk, and further caused the monitor to be sampled near the same side. Note that the most likely location is aligned with the keyboard. In the second scene, most skeletons are sitting on the couch or standing in front of it, resulting a dense distribution near the TV stand. The third scene has two major areas—near the chairs and near the sofa.

Table 1 gives quantitative evaluation of our algorithms based on two metrics: *location difference*, measuring the Euclidean distance between the predicted and the labeled locations, and also *height difference*.

In the task of placing new objects, using object context ('obj') beat other baseline methods, especially for the laptop, monitor, keyboard and mouse types. This was because of the strong spatial relationships among these objects. However, our method based on human context ('DP') still outperformed the object context. By considering human poses, it improved the placement

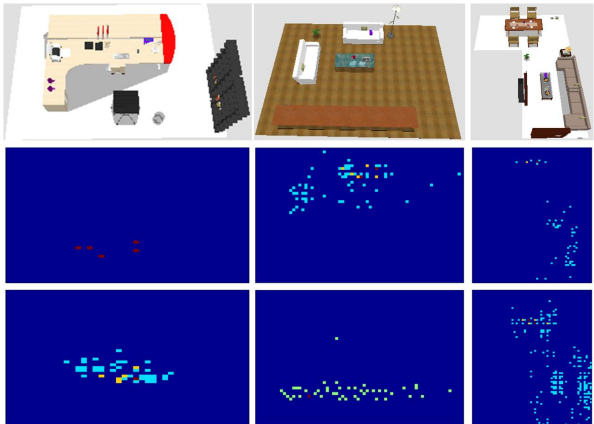

*Figure 6:* Given a scene (first row), our algorithm samples human poses (second row) and object placements (third row), for placing a monitor, TV and book respectively. (Red color means higher frequency, blue zero.)

of objects that have weaker connection to others, such as book, TV, decoration and shoes.

The task of arranging objects in an empty room is quite challenging since there is no object context for the first few placements. Not surprisingly, we found that the object-context method performed poorly, even worse than the simple height-preference rule. The performance of our methods was also affected. However, the sampled human poses could still pick up hints from the furniture in the scene, using room geometry (with no semantic labels). Our experiments also showed that the finite mixture model using human context performed better than other baselines, but not as well as the ones using DPs.

In both tasks, our human-context algorithm successfully predicted object placements within 1.6 meters on average. The average error in height was only 0.1 meters. By combining human- and object-context, the error was further reduced—indicating that they provide some complementary context.

**Results on real scenes.** To demonstrate that our algorithm is also robust in real scenes, we tested on point clouds taken from five real offices/apartments from the dataset published in Jiang et al. (2012a). Similar to Jiang et. al., we evaluated the final arrangements by asking two human subjects (one male and one female, not associated with the project) to label the placements for each object as semantically correct or not, and also score the overall object arrangement on a scale from 0 to 5.

Table 2 shows the results on the five real scenes. The office2 scene has only one big table in the room, therefore the number of semantically correct placements ('Co') is 100% for every method. The open-area method performed well in this scene because it placed objects spread around the table, unlike some baselines, which piled up objects. However, our and Jiang et al.'s approach both arranged objects more meaningfully, i.e., books were stacked together, while a keyboard, laptop and mouse were placed close to each other. The Apt2 scene has many different layers for placing and thus some baselines could not identify semantically correct placing areas for objects. The DP, however, performed much better by, for example, placing shoes at the bottom level of a shelf, while food and books are on the middle level or on a table. Jiang et al.'s approach sometimes put the laptop on a shelf making it difficult for human to access.

**Results on robotic placements.** Finally, given the predicted arrangement of a scene, we used our PO-LAR and Kodiak PR2 robots to place the objects in simulation. The experimental details and results are provided in Jiang & Saxena (2012).

## 7. Conclusion

In this work, we applied human context to the task of arranging objects in a 3D scene. The key idea was that human poses and object placements relate strongly to each other in terms of object affordances, access effort and activity relevance. We designed potential functions based on spatial features to capture these relations. In an unsupervised learning setting where we are given a collection of 3D scenes containing objects, we learned the distribution of human poses as well as object placements using a variant of Dirichlet process. Our extensive experiments on 20 different rooms with 19 types of object showed that the arrangements are improved by considering human context. We also tested this on a personal robot in arranging and placing items.

## Acknowledgements

This research was funded by Microsoft Faculty Fellowship and Alfred P. Sloan Fellowship to Saxena.

## References

Dreyfuss, H. *Designing for People.* Allworth Press, 1955.

Edsinger, A. and Kemp, C.C. Manipulation in human environments. In *Humanoid Robots*, 2006.

Felzenszwalb, P., McAllester, D., and Ramanan, D. A discriminatively trained, multiscale, deformable part model. In *CVPR*, 2008.

Fisher, M., Savva, M., and Hanrahan, P. Characterizing structural relationships in scenes using graph kernels. In *SIGGRAPH*, 2011.

Grabner, H., Gall, J., and Gool, L. J. V. What makes a chair a chair? In *CVPR*, 2011.

Gupta, A., Satkin, S., Efros, A.A., and Hebert, M. From 3D scene geometry to human workspace. In *CVPR*, 2011.

# Learning Object Arrangements in 3D Scenes using Human Context

*Table 1:* Results of arranging new objects (top) and arranging empty rooms (bottom). The last column also shows the standard error.

| | | book | clean tool | lap-top | mon-itor | key-board | mouse | pen | deco | dish-ware | pan | cush-ion | TV | desk light | floor light | uten-sil | food | shoe | re-mote | phone | AVG |
|---|---|---|---|---|---|---|---|---|---|---|---|---|---|---|---|---|---|---|---|---|---|
| | | | | | | | | *Arranging new objects* | | | | | | | | | | | | | |
| location (m) | open | 3.09 | 3.58 | 2.70 | 2.22 | 2.08 | 3.30 | 3.45 | 3.11 | 2.34 | 3.64 | 4.27 | 4.12 | 3.55 | 2.93 | 2.55 | 2.86 | 1.71 | 3.04 | 3.42 | 3.05±.12 |
| | height | 2.15 | 1.72 | 1.63 | 0.79 | 0.76 | 1.47 | 1.96 | 1.93 | 1.59 | 1.84 | 2.35 | 3.77 | 1.49 | 3.31 | 1.48 | 2.05 | 1.31 | 1.63 | 1.76 | 1.81±.17 |
| | room | 2.05 | 2.42 | 1.39 | 1.08 | 0.91 | 1.76 | 1.55 | 2.52 | 2.43 | 1.73 | 2.43 | 3.80 | 1.79 | 2.68 | 2.07 | 2.30 | 1.29 | 2.22 | 1.70 | 2.00±.14 |
| | obj. | 1.74 | 1.83 | 1.19 | 0.56 | 0.61 | 1.02 | 0.58 | 2.01 | 1.75 | 1.74 | 3.22 | 3.61 | 1.48 | 2.83 | 1.88 | 1.29 | 1.97 | 1.99 | 1.22 | 1.71±.23 |
| | class. | 3.22 | 2.46 | 3.06 | 2.06 | 2.31 | 3.01 | 2.73 | 3.26 | 2.42 | 4.29 | 4.70 | 5.18 | 2.85 | 2.76 | 2.62 | 1.91 | 2.12 | 3.23 | 3.22 | 3.02±.18 |
| | FMM | 1.84 | 1.65 | 0.91 | 0.80 | 0.69 | 1.19 | 0.93 | 2.01 | 1.64 | 1.86 | 2.38 | 3.01 | 1.41 | 2.90 | 1.51 | 1.60 | 1.44 | 1.32 | 1.15 | 1.59±.17 |
| | DP | 1.73 | 1.67 | 1.06 | 0.77 | 0.76 | 1.05 | 0.71 | 2.06 | 1.35 | 1.88 | 2.08 | 2.78 | 1.23 | 2.71 | 1.22 | 1.23 | 1.48 | 0.94 | 1.47 | 1.48±.18 |
| | DP+obj | 1.63 | 1.71 | 1.03 | 0.71 | 0.74 | 1.00 | 0.72 | 1.86 | 1.15 | 2.09 | 1.90 | 2.53 | 1.13 | 2.50 | 1.17 | 1.13 | 1.73 | 1.38 | 1.19 | 1.44±.18 |
| height (m) | open | 0.56 | 0.65 | 0.51 | 0.38 | 0.38 | 0.66 | 0.70 | 0.21 | 0.76 | 0.84 | 0.50 | 0.55 | 0.62 | 0.00 | 0.88 | 0.58 | 0.00 | 0.35 | 0.69 | 0.52±.06 |
| | height | 0.17 | 0.14 | 0.11 | 0.04 | 0.04 | 0.12 | 0.21 | 0.22 | 0.22 | 0.14 | 0.18 | 0.10 | 0.00 | 0.17 | 0.10 | 0.00 | 0.00 | 0.19 | 0.10 | 0.12±.02 |
| | room | 0.52 | 0.86 | 0.34 | 0.18 | 0.18 | 0.41 | 0.35 | 0.94 | 0.81 | 1.19 | 0.55 | 0.52 | 0.43 | 0.00 | 0.07 | 0.00 | 0.44 | 0.36 | 0.53±.08 | |
| | obj. | 0.13 | 0.19 | 0.09 | 0.02 | 0.03 | 0.03 | 0.27 | 0.16 | 0.16 | 0.24 | 0.39 | 0.11 | 0.07 | 0.22 | 0.17 | 0.00 | 0.15 | 0.03 | 0.13±.02 | |
| | class. | 0.46 | 0.27 | 0.49 | 0.38 | 0.38 | 0.53 | 0.68 | 0.34 | 0.42 | 0.84 | 0.48 | 0.48 | 0.58 | 0.46 | 0.36 | 0.26 | 0.29 | 0.34 | 0.60 | 0.46±.03 |
| | FMM | 0.12 | 0.08 | 0.08 | 0.05 | 0.03 | 0.07 | 0.03 | 0.29 | 0.15 | 0.20 | 0.14 | 0.17 | 0.12 | 0.42 | 0.19 | 0.54 | 0.19 | 0.09 | 0.05 | 0.16±.03 |
| | DP | 0.17 | 0.08 | 0.07 | 0.05 | 0.05 | 0.12 | 0.08 | 0.22 | 0.17 | 0.22 | 0.10 | 0.22 | 0.09 | 0.01 | 0.16 | 0.10 | 0.01 | 0.12 | 0.10 | 0.11±.01 |
| | DP+obj | 0.14 | 0.08 | 0.04 | 0.03 | 0.02 | 0.04 | 0.03 | 0.24 | 0.15 | 0.18 | 0.14 | 0.13 | 0.06 | 0.02 | 0.15 | 0.05 | 0.01 | 0.08 | 0.03 | 0.09±.01 |
| | | | | | | | | *Arranging empty rooms* | | | | | | | | | | | | | |
| location (m) | open | 2.47 | 2.09 | 1.69 | 1.67 | 1.23 | 2.07 | 2.22 | 2.19 | 2.03 | 2.20 | 2.75 | 4.24 | 2.72 | 3.51 | 1.78 | 2.10 | 2.04 | 1.68 | 2.57 | 2.28±.20 |
| | height | 2.12 | 1.78 | 1.60 | 0.63 | 0.94 | 1.52 | 1.41 | 1.96 | 1.69 | 1.78 | 2.30 | 3.73 | 1.58 | 3.54 | 1.51 | 2.01 | 1.31 | 1.63 | 1.66 | 1.84±.16 |
| | room | 2.04 | 2.56 | 1.35 | 1.16 | 1.14 | 1.64 | 1.71 | 2.56 | 2.38 | 1.61 | 2.30 | 3.82 | 1.82 | 2.68 | 2.00 | 2.01 | 1.29 | 2.35 | 1.64 | 2.00±.14 |
| | obj. | 2.62 | 2.26 | 2.02 | 1.49 | 1.50 | 2.34 | 2.21 | 2.73 | 2.12 | 1.75 | 4.28 | 3.46 | 2.44 | 3.27 | 1.71 | 1.49 | 1.87 | 2.71 | 2.07 | 2.33±.17 |
| | class. | 2.60 | 3.24 | 2.79 | 1.94 | 2.23 | 2.85 | 2.25 | 2.65 | 2.77 | 3.47 | 4.24 | 4.74 | 1.99 | 3.13 | 2.72 | 2.06 | 2.73 | 3.05 | 3.29 | 2.88±.23 |
| | FMM | 2.14 | 1.66 | 1.22 | 1.17 | 1.01 | 1.42 | 1.52 | 2.11 | 1.72 | 1.85 | 3.08 | 3.09 | 1.70 | 2.02 | 1.57 | 1.73 | 1.09 | 1.40 | 1.52 | 1.74±.11 |
| | DP | 1.78 | 1.57 | 1.34 | 1.16 | 0.91 | 1.53 | 0.96 | 1.96 | 1.50 | 1.58 | 2.59 | 2.86 | 1.27 | 2.76 | 1.66 | 1.27 | 1.89 | 1.39 | 1.43 | 1.65±.20 |
| | DP+obj | 1.65 | 1.74 | 1.12 | 0.59 | 0.72 | 1.15 | 0.82 | 2.06 | 1.19 | 2.21 | 3.17 | 3.40 | 1.24 | 3.15 | 1.32 | 1.47 | 1.38 | 1.61 | 1.03 | 1.63±.19 |
| height (m) | open | 0.47 | 0.55 | 0.42 | 0.37 | 0.33 | 0.43 | 0.58 | 0.32 | 0.65 | 0.73 | 0.44 | 0.53 | 0.49 | 0.00 | 0.59 | 0.52 | 0.04 | 0.18 | 0.62 | 0.43±.04 |
| | height | 0.17 | 0.11 | 0.11 | 0.04 | 0.04 | 0.13 | 0.11 | 0.25 | 0.22 | 0.23 | 0.14 | 0.14 | 0.00 | 0.00 | 0.17 | 0.00 | 0.00 | 0.03 | 0.13 | 0.12±.01 |
| | room | 0.49 | 0.81 | 0.28 | 0.17 | 0.15 | 0.33 | 0.28 | 0.88 | 0.79 | 1.09 | 0.55 | 0.52 | 0.40 | 0.00 | 1.65 | 0.77 | 0.00 | 0.43 | 0.28 | 0.49±.07 |
| | obj. | 0.47 | 0.56 | 0.47 | 0.31 | 0.33 | 0.57 | 0.60 | 0.24 | 0.64 | 0.71 | 0.42 | 0.41 | 0.55 | 0.04 | 0.68 | 0.47 | 0.00 | 0.31 | 0.61 | 0.44±.04 |
| | LR | 0.48 | 0.70 | 0.50 | 0.38 | 0.38 | 0.64 | 0.43 | 0.39 | 0.68 | 0.79 | 0.44 | 0.62 | 0.30 | 0.01 | 0.59 | 0.46 | 0.29 | 0.25 | 0.56 | 0.47±.04 |
| | FMM | 0.19 | 0.14 | 0.11 | 0.10 | 0.07 | 0.14 | 0.14 | 0.20 | 0.23 | 0.20 | 0.13 | 0.16 | 0.22 | 0.44 | 0.32 | 0.56 | 0.19 | 0.12 | 0.15 | 0.20±.03 |
| | DP | 0.15 | 0.11 | 0.08 | 0.09 | 0.07 | 0.15 | 0.06 | 0.22 | 0.22 | 0.23 | 0.13 | 0.23 | 0.03 | 0.03 | 0.22 | 0.09 | 0.02 | 0.08 | 0.06 | 0.12±.01 |
| | DP+obj | 0.14 | 0.12 | 0.05 | 0.04 | 0.07 | 0.10 | 0.05 | 0.19 | 0.19 | 0.21 | 0.14 | 0.24 | 0.06 | 0.06 | 0.19 | 0.12 | 0.00 | 0.08 | 0.07 | 0.11±.01 |

*Table 2:* Results on arranging five real point-cloud scenes (3 offices & 2 apartments). The number of objects for placing are 4, 18, 18, 21 and 18 in each scene respectively. **Co:** % of semantically correct placements, **Sc:** average score (0-5).

| | office1 | | office2 | | office3 | | apt1 | | apt2 | | Average | |
|---|---|---|---|---|---|---|---|---|---|---|---|---|
| | Co | Sc | Co | Sc | Co | Sc | Co | Sc | Co | Sc | Co | Sc |
| open | 100 | 2.5 | 100 | 3.5 | 30 | 1.5 | 63 | 2.5 | 45 | 3.0 | 68 | 2.6 |
| height | 80 | 3.0 | 100 | 2.0 | 60 | 2.5 | 50 | 2.5 | 75 | 3.5 | 73 | 2.7 |
| room | 100 | 4.0 | 100 | 2.5 | 0 | 0.5 | 20 | 0.5 | 35 | 1.5 | 51 | 1.8 |
| obj. | 100 | 4.5 | 100 | 3.0 | 45 | 1.0 | 20 | 1.5 | 78 | 3.3 | 68 | 2.7 |
| class. | 100 | 3.5 | 100 | 2.0 | 20 | 0.5 | 30 | 2.0 | 33 | 2.0 | 57 | 2.0 |
| Jiang12 | 100 | 4.5 | 100 | 4.2 | 87 | 3.5 | 65 | 3.2 | 75 | 3.0 | 85 | 3.7 |
| FMM | 100 | 3.5 | 100 | 2.0 | 83 | 3.8 | 63 | 3.5 | 63 | 3.0 | 82 | 3.2 |
| DP | 100 | **5.0** | 100 | 4.3 | 91 | 4.0 | 74 | 3.5 | **88** | **4.3** | 90 | 4.2 |
| DP+obj | 100 | 4.8 | 100 | **4.5** | **92** | **4.5** | **89** | **4.1** | 81 | 3.5 | **92** | 4.2 |

Heitz, G. and Koller, D. Learning spatial context: Using stuff to find things. In *ECCV*, 2008.

Heitz, G., Gould, S., Saxena, A., and Koller, D. Cascaded classification models: Combining models for holistic scene understanding. In *NIPS*, 2008.

Hoiem, D., Efros, A.A., and Hebert, M. Putting objects in perspective. In *CVPR*, 2006.

Jain, D., Mosenlechner, L., and Beetz, M. Equipping robot control programs with first-order probabilistic reasoning capabilities. In *ICRA*, 2009.

Jiang, Y. and Saxena, A. Hallucinating humans for learning robotic placement of objects. In *ISER*, 2012.

Jiang, Y., Lim, M., Zheng, C., and Saxena, A. Learning to place new objects in a scene. *IJRR*, 2012a.

Jiang, Y., Zheng, C., Lim, M., and Saxena, A. Learning to place new objects. In *ICRA*, 2012b.

Koppula, H.S., Anand, A., Joachims, T., and Saxena, A. Semantic labeling of 3D point clouds for indoor scenes. In *NIPS*, 2011.

Lee, D.C., Gupta, A., Hebert, M., and Kanade, T. Estimating spatial layout of rooms using volumetric reasoning about objects and surfaces. In *NIPS*, 2010.

Lee, M. W. and Cohen, I. A model-based approach for estimating human 3D poses in static images. *IEEE Trans PAMI*, 28:905–916, 2006.

Li, C., Kowdle, A., Saxena, A., and Chen, T. Towards holistic scene understanding: Feedback enabled cascaded classification models. In *NIPS*, 2010.

Ly, D., Saxena, A., and Lipson, H. Co-evolutionary predictors for kinematic pose inference from rgbd images. In *GECCO*, 2012.

Neal, R.M. Markov chain sampling methods for dirichlet process mixture models. *Journal of computational and graphical statistics*, pp. 249–265, 2000.

Saxena, A., Chung, S.H., and Ng, A. Learning depth from single monocular images. In *NIPS*, 2005.

Schuster, M., Okerman, J., Nguyen, H., Rehg, J., and Kemp, C. Perceiving clutter and surfaces for object placement in indoor environments. In *Humanoid Robots*, 2010.

Sung, J., Ponce, C., Selman, B., and Saxena, A. Human activity detection from RGBD images. In *ICRA*, 2012.

Teh, Y. W. Dirichlet process. *Encyclopedia of Machine Learning*, pp. 280–287, 2010.